\documentclass[10pt,twocolumn,letterpaper]{article}

\usepackage{iccv}
\usepackage{times}
\usepackage{epsfig}
\usepackage{amsmath}
\usepackage{amssymb}
\usepackage{subcaption}
\usepackage{mathabx}
\usepackage{comment}
\usepackage{amsfonts}
\usepackage{booktabs}
\usepackage{graphicx}
\usepackage{multirow}

\usepackage{capt-of}
\usepackage{pifont}
\newcommand{\xmark}{\ding{55}}%
\usepackage{tabularx,colortbl}
\definecolor{lavendergray}{rgb}{0.77, 0.76, 0.82}

\usepackage{floatrow}
\newfloatcommand{capbtabbox}{table}[][\FBwidth]

\usepackage[numbers,sort]{natbib}

\usepackage[pagebackref=true,breaklinks=true,letterpaper=true,colorlinks,bookmarks=false]{hyperref}
\usepackage[dvipsnames]{xcolor}

\newcommand{\weidi}[1]{{\textcolor{blue}{[Weidi: #1]}}}

\newcommand{\erika}[1]{{\textcolor{cyan}{[Erika: #1]}}}

\iccvfinalcopy 

\def\iccvPaperID{2132} 

\begin{document}
\ificcvfinal\pagestyle{empty}\fi

\title{Self-supervised Video Object Segmentation by Motion Grouping}

\title{Self-supervised Video Object Segmentation by Motion Grouping}

\author{Charig Yang \qquad \quad Hala Lamdouar \quad \qquad Erika Lu \quad \qquad  
Andrew Zisserman \qquad \quad Weidi Xie  \\ [2pt]
Visual Geometry Group, University of Oxford \\
{\tt\small \{charig,lamdouar,erika,az,weidi\}@robots.ox.ac.uk}\\
\url{https://charigyang.github.io/motiongroup/}}

\ificcvfinal\thispagestyle{empty}\fi

\twocolumn[{%
\renewcommand\twocolumn[1][]{#1}%
\maketitle
\begin{center}
    \centering
    \includegraphics[width=\textwidth]{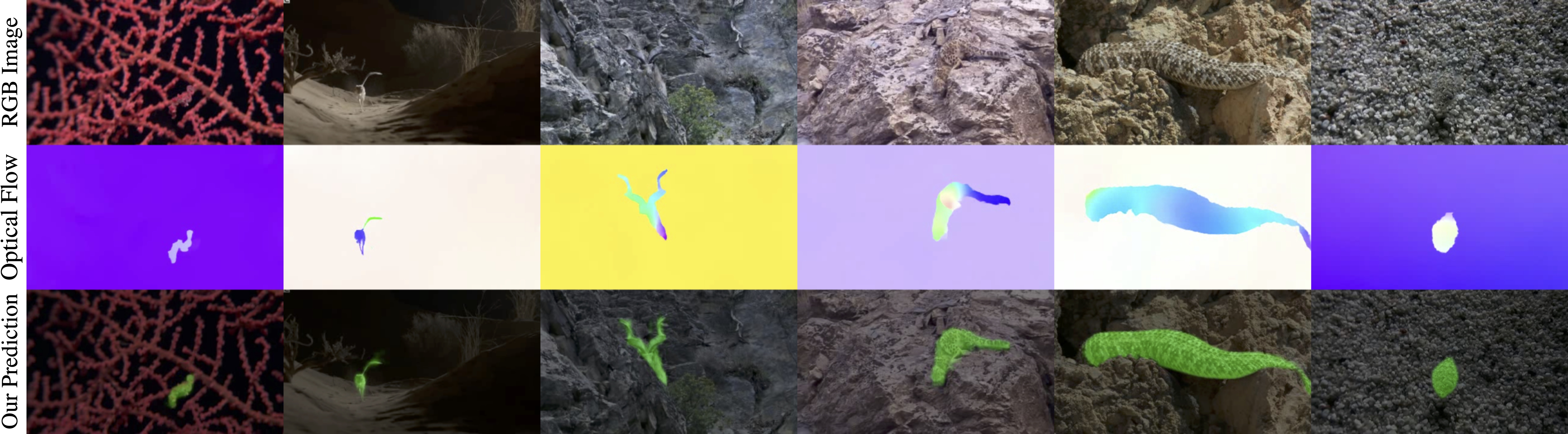}
    \captionof{figure}{ 
    \textbf{Segmenting camouflaged animals.}
    Motion plays a critical role in augmenting the capability of our visual system for perceptual grouping in complex scenes -- for example, in these sequences (MoCA dataset~\cite{Lamdouar20}), 
    the visual appearance~(RGB images) is clearly uninformative. 
    In this paper, we propose a self-supervised approach to segment objects using \textit{only} motion, 
    \ie~optical flow.
    From top to bottom rows, 
    we show the video frames, optical flow between consecutive frames, 
    and the segmentation produced by our approach.}
    \label{fig:teaser}
    \vspace{0.5 cm}
\end{center}%
}]

\begin{abstract}
\pagenumbering{gobble}
Animals have evolved highly functional visual systems to understand motion,
assisting perception even under complex environments.
In this paper, 
we work towards developing a computer vision system able
to segment objects by exploiting motion cues,
\ie~motion segmentation.
To achieve this,
we introduce a simple variant of the Transformer to segment optical flow frames into primary objects and the background, which can be trained in a self-supervised manner, \ie~without using any manual annotations.
Despite using only optical flow as input,
our approach achieves superior results compared to previous state-of-the-art self-supervised methods on public benchmarks~(DAVIS2016, SegTrackv2, and FBMS59),
while being an order of magnitude faster. 
We additionally evaluate on a challenging camouflage dataset~(MoCA),
significantly outperforming the other self-supervised approaches, and competitive to the top supervised approach,
highlighting the importance of motion cues,
and the potential bias towards appearance in existing video segmentation models.
\end{abstract}
\vspace{-1.4cm}

\section{Introduction}


When looking around the world, 
we effortlessly perceive a complex scene as a set of distinct objects.
This phenomenon is referred to as \emph{perceptual grouping} --
the process of organizing the incoming visual information --
and is usually considered a fundamental cognitive ability that enables understanding and interacting  
with the world efficiently.
How do we accomplish such a remarkable perceptual achievement, given that the visual input is, 
in a sense, just a spatial distribution of variously colored individual points/pixels? 
In 1923, 
Wertheimer~\cite{Wertheimer1923n} first introduced the \emph{Gestalt principles}
with the goal of formulating the underlying causes by which sensory data is organized into groups, 
or Gestalten. 
The principles are much like heuristics with ``a bag of tricks'' \cite{ramachandran1985guest} that the visual system may exploit for grouping, for example, proximity, similarity, closure, continuation, common fate, \etc.

In computer vision, 
\emph{perceptual grouping} is often closely related to the problem of segmentation, 
\ie~extracting the objects with arbitrary shape~(pixel-wise labels) from cluttered scenes.
In the recent literature of semantic or instance segmentation,
tremendous progress has been made by training deep neural networks on image or video datasets.
While 
it is exciting to see machines with the ability to detect, 
segment, and classify objects in images or video frames,
training such segmentation models through supervised learning  requires massive human annotation, and consequently
limiting their scalability.
Even more importantly, 
the assumption that objects can be well-identified by their appearance alone in static frames
is often an oversimplification --
objects are not always visually distinguishable from their background environment.
For instance when trying to discover camouflaged animals/objects from the background~(Figure~\ref{fig:teaser}),
extra cues, such as motion or sound, are usually required.

Among the numerous cues, 
motion is usually simple to obtain as it can be implicitly generated from unlabeled videos.
In this paper, we aim to exploit such cues for object segmentation in a self-supervised manner, 
\ie~\emph{zero} human annotation is required for training.
At a high level, we aim to exploit the common fate principle, 
with the basic assumption being that
\textbf{elements tend to be perceived as a group if they move in the same direction at the same rate (have similar optical flow)}.
Specifically,
we tackle the problem by training a generative model that decomposes the optical flow into foreground (object) and background layers,
describing each as a homogeneous field, with discontinuities occurring only between layers.
We adopt a variant of the Transformer~\cite{Vaswani17},
with the self-attention being replaced by slot attention~\cite{locatello2020object},
where iterative grouping and binding have been built into the architecture.
With some critical architectural changes,
we show that pixels undergoing similar motion are grouped together and assigned to the same layer.


To summarize, we make the following contributions:
\emph{first}, 
we introduce a simple architecture for video object segmentation by exploiting motions, 
using only optical flow as input.
\emph{Second},
we propose a self-supervised proxy task that is used to train the architecture without any manual supervision.
To validate these contributions,
we conduct thorough ablation studies on the components that are key to the success of our architecture, 
such as a consistency loss on optical flow computed from various frame gaps. 
We evaluate the proposed architecture on public benchmarks 
(DAVIS2016~\cite{Perazzi2016}, SegTrackv2~\cite{FliICCV2013}, and FBMS59~\cite{Ochs14}), 
outperforming previous state-of-the-art self-supervised models,
with comparable performance to the supervised approaches.
Moreover, 
we also evaluate on a camouflage dataset~(MoCA~\cite{Lamdouar20}),
demonstrating a significant performance improvement over the other self- and supervised approaches,
highlighting the importance of motion cues, 
and the potential bias towards visual appearance in existing video segmentation models. 
\section{Related Work}
\par{\noindent \textbf{Video object segmentation}}
has been a longstanding task in computer vision,
which involves assigning pixels (or edges) of an image into groups (e.g. objects).
In recent literature~\cite{Bideau16a,Ponttuset17,Brox10,Ochs11,Papazoglou13,Jain17,Tokmakov19,Dave19,iccv19_stm,cvpr19_feelvos,Vondrick18,Wang19,Lai19,Lai20,tpami18_osvos-s,bmvc17_OnAVOS,cvpr17_OSVOS,Fragkiadaki12,Keuper15,Jun17,Fan19,Yang19,Ponttuset17,xu18,lu2020learning}, 
two protocols have attracted increasing interest from the vision community,
namely, semi-supervised video object segmentation~({\bf semi-supervised VOS}),
and unsupervised video object segmentation~({\bf unsupervised VOS}).
The former aims to re-localize one or multiple targets that are specified in the first frame of a video with pixel-wise masks, and the latter considers automatically segmenting the object of interest~(usually the most salient one) from the background in a video sequence.
Despite being called {\bf unsupervised VOS}, 
in practice,
the popular methods to address such problems extensively rely on supervised training, for example, 
by using two-stream networks~\cite{Papazoglou13,Jain17,Tokmakov19,Dave19} trained on large-scale external datasets.
As an alternative,
in this work, we consider a \emph{completely unsupervised} approach,
where no manual annotation is used for training whatsoever.\\[-10pt]

\par{\noindent \textbf{Motion segmentation}}
shares some similarity with unsupervised VOS, 
but focuses on discovering \emph{moving} objects.
In the literature,
\cite{Brox10,Ochs11,Keuper15,Sivic04a,xie19} consider clustering the pixels with similar motion patterns;
\cite{Tokmakov17,Tokmakov19,Dave19} train deep networks to map the motions to segmentation masks.
Another line of work has  tackled the problem by explicitly leveraging the independence of motion between the moving object and its background. 
For instance,~\cite{yang_loquercio_2019} proposes an adversarial setting, where a generator is trained to produce masks, altering the input flow, such that the inpainter fails to estimate the missing information. 
In~\cite{Bideau16,Bideau18,Lamdouar20}, 
the authors propose to highlight the independently moving object by compensating for the background motion,
either by registering consecutive frames, or explicitly estimating camera motion.
In constrained scenarios, such as autonomous driving, 
\cite{Ranjan19} proposes to jointly optimize depth, camera motion, optical flow and motion segmentation.\\[-10pt]

\par{\noindent \textbf{Optical flow}} 
computation is one of the fundamental tasks in computer vision. 
Deep learning methods allow efficient computation of optical flow, 
both in training on synthetic data~\cite{Sun2018PWC-Net,Teed20}, 
or learning with photometric loss in self-supervised \cite{liu2020learning, Liu:2019:SelFlow} setting. 
In practise, flow has been useful for a wide range of problems, 
for example, 
pose estimation \cite{doersch2019sim2real}, 
representation learning \cite{Mahendran18,Han20b}, 
segmentation \cite{Brox10},
and occasionally even used in lieu of appearance cues (RGB images) for
tracking \cite{sidenbladh2000stochastic}.\\[-10pt]

\par{\noindent \textbf{Transformer architectures}} 
have proven extremely adept at modelling long-term relationships 
within an input sequence via attention mechanisms.
Originally used for language tasks~\cite{Vaswani17,Devlin2018,brown2020language}, 
they have since been adapted to solve popular computer vision problems 
such as image classification~\cite{dosovitskiy2020image}, 
generation~\cite{Chen_generative20,Ramesh_DALLE21},
video understanding~\cite{Wang2018,girdhar2019video,Bertasius21}, 
object detection~\cite{carion2020end},
and zero-shot classification~\cite{Radford_CLIP21}.
In this work, we take inspiration from a specific variant of self-attention,
namely slot attention~\cite{locatello2020object},
which was demonstrated to be effective for learning object-centric representations on synthetic data, \eg~CLEVR \cite{johnson2017clevr}.\\[-10pt]

\par{\noindent \textbf{Layered representations}} 
were originally proposed by Wang and Adelson~\cite{Wang94}
to represent a video as a composition of layers with simpler motions.
Since then, layered representations have been widely adopted in computer vision~\cite{brostow1999motion,Jojic01,zitnick2004high,Kumar08,Xue15},
often to estimate optical flow~\cite{Sun12,Sun13,Wulff14,Wulff15}.
More recently, deep learning-based layer decomposition methods have been 
used to infer depth for novel view synthesis~\cite{Zhou18,Srinivasan19}, separate reflections and other semi-transparent effects~\cite{Alayrac19a,Alayrac19b,Gandelsman19,lu2020}, or perform foreground/background estimation~\cite{Gandelsman19}.
These works operate on RGB inputs and produce RGB layers, whereas we propose a layered decomposition of optical flow inputs for unsupervised moving object discovery.\\[-10pt]

%

\par{\noindent \textbf{Object-centric representations}}
interpret scenes with ``objects'' as the basic building blocks~(instead of individual pixels), 
which is considered an essential step towards human-level generalization.
There is a rich literature on this topic,
for example, IODINE \cite{Greff19} uses iterative variational inference to infer a set of latent variables recurrently, 
with each representing one object in an image.
Similarly, MONet~\cite{Burgess19} and GENESIS~\cite{Engelcke20} also adopt multiple encoding-decoding steps.
In contrast, \cite{locatello2020object} proposes Slot Attention,
which enables single step encoding-decoding with iterative attention.
However, all works mentioned above have only shown applications for synthetic datasets, \eg~CLEVR \cite{johnson2017clevr}. 
In this paper, we are the first to demonstrate its use for object segmentation of realistic videos
by exploiting motion, where the challenging nuances in visual appearance
(\eg the complex background textures) have been removed.



\begin{figure*}[!htb]
\includegraphics[width=.99\textwidth]{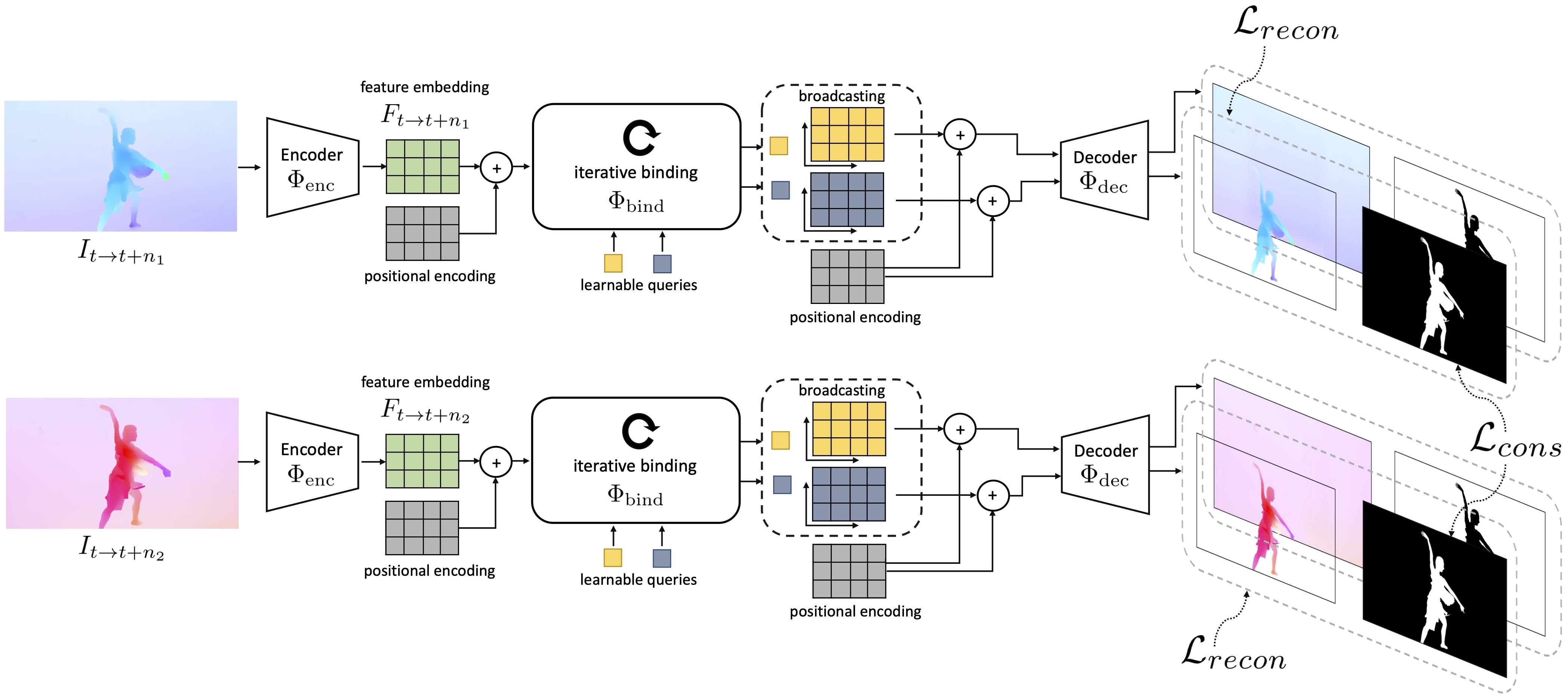}
\centering
\vspace{-.45cm}
\caption{ \small
\textbf{Pipeline.}
Our model takes optical flow $I_{t\rightarrow t+n}$ as input, 
and outputs a set of reconstruction and opacity layers.
Specifically, it consists of three components: feature encoding, 
iterative binding, and decoding to layers, which are combined to reconstruct the input flow.
To resolve motion ambiguities~(small motion), or noise in optical flow,
consistency between two flow fields computed under different frame gaps is enforced during training.
At inference time, only the top half of the figure is used to predict masks from a single-step flow.
}
\vspace{-.1cm}
\label{fig:arch}
\end{figure*}

\vspace{-.1cm}
\section{Method}
\vspace{-.1cm}
\label{sec:method}



Our goal is to take an input optical flow frame and predict a segment containing the moving object.
We propose to train this model in a self-supervised manner,
with an autoencoder-type framework.
Specifically, our model outputs two layers: one representing the background, and the other for one or more moving objects in the foreground, as well as their oppacity layers~(weighted masks).
Formally:
\begin{align}
     \{\hat{I}_{t \rightarrow t+n}^i, \alpha_{t \rightarrow t+n}^i\}_{i=1}^N = \Phi(I_{t \rightarrow t+n}) 
\end{align}
where $I_{t\rightarrow t+n}$ refers to the $t$ to $t+n$ input flow~(backward flow when $n<0$),
$\Phi(\cdot)$ is the parametrized model, 
$\hat{I}_{t \rightarrow t+n}^i$ is the $i$th layer reconstruction, $\alpha_{t \rightarrow t+n}^i$ is its mask,
and $N=2$ is the number of layers (foreground and background).
These layers can then be composited linearly to reconstruct the input image $I_{t \rightarrow t+n}$: 
\begin{align}
     \hat{I}_{t \rightarrow t+n} = \sum_{i=1}^{N} \alpha_{t \rightarrow t+n}^i \hat{I}_{t \rightarrow t+n}^i
     \label{eq:layers}
 \end{align}


\subsection{Flow Segmentation Architecture}
\label{sec:arch}
For simplicity, we first consider the case of a single flow field as input~(depicted in the top part of Figure~\ref{fig:arch}).
The entire model consists of three components: 
(1) a CNN encoder to extract a compact feature representation, 
(2) an iterative binding module with learnable queries that plays a similar role as soft clustering, 
\ie~assigning each pixel to one of motion groups,
and (3) a CNN decoder that individually decodes each query to full resolution layer outputs (where thresholding the alpha channel yields the predicted segment).\\[-6pt]

\par{\noindent \textbf{CNN encoder.}}
We first pass the precomputed optical flow between two frames,
$I_{t \rightarrow t+n} \in \mathcal{R}^{3 \times H_0 \times W_0}$,
to a CNN encoder ~$\Phi_{\text{enc}}$,
which outputs a lower-resolution feature map:
\begin{align}
    F_{t \rightarrow t+n} = \Phi_{\text{enc}}(I_{t \rightarrow t+n}) \in \mathcal{R}^{D \times H \times W}
\end{align}
where $H_0, W_0$ and $H, W$ refer to the spatial dimensions of the input and output feature maps respectively. 
Note that, we convert the flow into a three-channel image 
using the traditional method in the optical flow literature~\cite{Sun2018PWC-Net}.\\[-6pt]

\par{\noindent \textbf{Iterative binding.}}
The iterative binding module ~$\Phi_{\text{bind}}$ aims to group image regions into single entities based on their similarities in motion,
\ie~pixels moving in the same direction at the same rate should be grouped together.
Intuitively, such a binding process requires a data-dependent parameter updating mechanism, 
iteratively enriching the model, gradually including more pixels undergoing similar motions.

To accomplish this task, we adopt a simple variant of slot attention~\cite{locatello2020object},
where instead of Gaussian-initialized slots, we use learnable query vectors. 
Slot attention has recently shown remarkable performance for object-centric representation learning, 
where the query vectors compete to explain parts of the inputs 
via a softmax-based attention mechanism,
and the representations of these slots are iteratively updated with a recurrent update function.
In our case of motion segmentation,
ideally, 
the final representation in each query vector separately encodes the moving object or the background,
which can then be decoded and combined to reconstruct the input flow fields.

Formally, our inputs to $\Phi_{\text{bind}}$ are
feature maps $F_{t \rightarrow t+n}$
and two learnable queries (representing foreground and background) $Q \in \mathcal{R}^{D \times 2}$. 
Learnable spatial positional encodings are summed with $F_{t \rightarrow t+n}$;
with some abuse of notation, we still refer to this sum as $F_{t \rightarrow t+n}$.
We use \emph{three} different linear transformations to generate the $query$ , $key$ and $value$: $q \in \mathcal{R}^{D \times 2}$, $k, v \in \mathcal{R}^{D \times HW}$,
\begin{align}
    q, \text{\hspace{3pt}} k, \text{\hspace{3pt}} v =  
    W^Q {\cdot} Q, \text{\hspace{3pt}}  W^K {\cdot} F_{t \rightarrow t+n}, 
    \text{\hspace{3pt}}  W^V {\cdot} F_{t \rightarrow t+n}
\end{align}
where $W^Q, W^K, W^V \in \mathcal{R}^{D \times D}$.

In contrast to the standard Transformer~\cite{Vaswani17},
the coefficients in slot attention are normalized over all slots.
This choice of normalization introduces competition between the slots to explain parts of the input,
and ensures each pixel is assigned to a query vector:
\begin{align}
    &\text{attn}_{i,j} := \frac{e^{M_{i,j}}}{\sum_l e^{M_{i,l}}} \text{\hspace{5pt}}  \\ \nonumber
    &M := \frac{1}{\sqrt{D}}k^T {\cdot} q, 
    \text{\hspace{5pt}attn} \in \mathcal{R}^{HW \times 2}
\end{align}
%
To aggregate the input values to their assigned query slot, 
a weighted mean is used as follows:
\begin{align}
    &U :=v {\cdot} A \text{\hspace{5pt}} \in \mathcal{R}^{D \times 2} \\ \nonumber
    &\text{where, \hspace{5pt}}A_{i,j} := \frac{\text{attn}_{i,j}}{\sum_{l} \text{attn}_{l,j}}
\end{align}
To maintain a smooth update of the query slots $Q$,
the aggregated vectors $U$ are fed into a recurrent function,
parametrized with Gated Recurrent Units~(GRU), 
\begin{align}
    Q := \text{GRU}(\text{inputs}=U, \text{states}=Q)
\end{align}
This whole binding process is then iterated $T$ times.
The pseudocode can be found in the Supplementary Material. \\[-9pt]


\par{\noindent \textbf{CNN decoder.}}
The CNN decoder ~$\Phi_{\text{dec}}$ individually decodes each of the slots to outputs of original resolution~($\{\hat{I}^i_{t \rightarrow t+n}, \alpha^i_{t \rightarrow t+n}\} \in \mathcal{R}^{4 \times H_0 \times W_0}$),
which includes an (unnormalized) single-channel alpha mask and the reconstructed flow fields.
Specifically, the input to the decoder is the slot vector broadcasted onto a 2D grid augmented with a learnable spatial positional encoding. \\ [-9pt]

\par{\noindent \textbf{Reconstruction. }}
Once each slot has been decoded,
we apply softmax to the alpha masks across the slot dimension, 
and use them as mixture weights to obtain the reconstruction $\hat{I}_{t \rightarrow t+n}$~(Eq.~\ref{eq:layers}).
Our reconstruction loss is an L2 loss between the input and reconstructed flow,
\begin{equation}
    \mathcal{L}_{recon} = \frac{1}{\Omega} \sum_{p \in \Omega} |I_{t \rightarrow t+n}(p) - \hat{I}_{t \rightarrow t+n}(p)|^2 
\end{equation}
where $p$ is the pixel index, and $\Omega$ is the entire spatial grid.\\[-8pt]

\par{\noindent \textbf{Entropy regularization. }}
We impose a pixel-wise entropy regularisation on inferred masks:
\begin{align}
    \mathcal{L}_{entr} = \frac{1}{\Omega} \sum_{p \in \Omega}(-\alpha^1_{t \rightarrow t+n}(p) \log \alpha^1_{t \rightarrow t+n}(p) \\ \nonumber
    -\alpha^0_{t \rightarrow t+n}(p) \log \alpha^0_{t \rightarrow t+n}(p))
\end{align}
This loss is zero when the alpha channels are one-hot, 
and maximum when they are of equal probability. 
Intuitively, this helps encourage the masks to be binary, 
which aligns with our goal in obtaining segmentation masks.\\[-8pt]


\par{\noindent \textbf{Instance normalisation.}}
In the case of motion segmentation, 
objects can only be detected if they undergo an independent motion from the camera;
thus previous work attempts to compensate for camera motion~\cite{Bideau16, Lamdouar20}. 
We are inspired by these ideas, but instead of explicitly estimating homography or camera motion, 
we take a poor-man's approach by simply using Instance Normalisation~(IN) \cite{ulyanov2016instance} 
in the CNN encoder and decoder, 
which normalizes each channel of the training sample independently.
Intuitively, the \emph{mean} activation tends to be dominated by the motions in the large homogeneous region, 
which is usually the background. 
This normalization, in combination with ReLU activations, 
helps gradually separate the background motion from the foreground motions.
This is experimentally shown in Section~\ref{sec:ablation}\\[-5pt]

\vspace{-.2cm}
\subsection{Self-supervised Temporal Consistency Loss}
\vspace{-.1cm}
\label{sec:consistency}
The segmentation computed for the current frame should be identical irrespective of whether the `second' frame is consecutive,
or earlier or later in time. We harness this constraint to form a self-supervised temporal consistency loss by first defining
a set of `second' frames, and then requiring consistency between their pairwise predictions. We describe the set first, followed by the loss.\\ [-5pt]

\par{\noindent \textbf{Multi-step flow. }}
As objects may be static for some frames, we make our predictions more robust by
leveraging observations from multiple timesteps.
We consider the flow fields computed from various temporal gaps as an input set, 
\ie~$\{I_{t \rightarrow t+n_1},I_{t \rightarrow t+n_2}\}$, $n_1, n_2 \in \{-2, -1, 1, 2\}$,
and use a permutation invariant consistency loss to encourage the model to predict 
the same foreground/background segmentation for all flow fields in the set.\\ [-5pt]

\par{\noindent \textbf{Consistency loss. }} 
We randomly sample two flow fields from the input set and pass them through the model~($\Phi(\cdot)$), 
outputting the flow reconstruction and alpha masks for each.
As the reconstruction loss is commutative, 
it is not guaranteed that the same slot will always output the background layer; 
therefore, we use a permutation-invariant consistency loss, 
\ie~only backpropagating through the lowest-error permutation:
\begin{align*}
    \mathcal{L}_{cons} =  \frac{1}{\Omega}
    \text{min}(\sum_{p \in \Omega} |\alpha^1_{t \rightarrow t+n_1}(p) - \alpha^1_{t \rightarrow t+n_2}(p)|^2, \\ 
    \sum_{p \in \Omega} |\alpha^1_{t \rightarrow t+n_1}(p) - \alpha^0_{t \rightarrow t+n_2}(p)|^2)
\end{align*}

Note that, this consistency enforcement only occurs during training. At inference time, a single-step flow is used, as shown in the top half of Figure~\ref{fig:arch}. \\ [-5pt]


\par{\noindent \textbf{Total loss.}}
The total loss for training the architecture is:
\begin{align}
    \mathcal{L}_{total} = \gamma_{r} \mathcal{L}_{recon} + \gamma_{c} \mathcal{L}_{cons} + \gamma_{e} \mathcal{L}_{entr}
\end{align}
we use $\gamma_r = 10^2$, $\gamma_c = 10^{-2}$ and $\gamma_e  = 10^{-2}$,
but we found the model to be fairly robust to these hyperparameters.
\subsection{Discussion}
\label{sec:discussion}
\par{\noindent \textbf{Differences from slot attention.}}
Slot attention was originally introduced
for self-supervised object segmentation for RGB images~\cite{locatello2020object}, 
and its usefulness was demonstrated on synthetic data (CLEVR~\cite{johnson2017clevr}),
where objects are made of primitive shapes with simple textures.
However, 
this assumption is unlikely to hold in the case of natural images or videos,
making it challenging to generalize such object-centric representations.

In this work, 
we build on the insight that although objects in images may not be naturally textureless, 
their motions typically are.
Hence, we develop the self-supervised object segmentation model by exploiting their optical flows, 
where the nuance in visual appearance is discarded, 
thus not restricted to simple synthetic cases.
As an initial trial, we experimented with the same setting as~\cite{locatello2020object}, 
where query vectors are sampled from a Gaussian distribution; 
however, we were unable to train it.
Instead, we use learnable embeddings here,
which we highlight as one of the architectural changes
critical to our model's success.
Other critical changes include instance normalization and temporal consistency, which we demonstrate in ablations in Section~\ref{sec:ablation}.\\ [-9pt]

\par{\noindent \textbf{Why does it work for motion segmentation?}}
The proposed idea can be seen as training a generative model to segment the flow fields.
With the layered formulation, 
reconstruction is limited to be a simple \emph{linear} composition of layer-wise flow,
decoded from a single slot vector.

Conceptually, this design has effectively introduced a representational bottleneck,
encouraging each slot vector to represent minimal information, \ie~homogeneous motion,
and with minimal redundancy~(mutual information) between slots.
All these properties make such an architecture well-suited to the task of segmenting objects undergoing independent motions.

\section{Experimental Setup}
\label{sec:experiment}

\subsection{Datasets}

\par{\noindent \textbf{DAVIS2016~\cite{Perazzi2016}}}
contains a total of 50 sequences ~(30 for training and 20 for validation), 
depicting diverse moving objects such as animals, people, and cars.
The dataset contains 3455 1080p frames with pixel-wise annotations at 480p for the predominantly moving object. \\[-8pt]

\par{\noindent \textbf{SegTrackv2~\cite{FliICCV2013}}} 
contains 14 sequences and 976 annotated frames. Each sequence contains 1-6 moving objects, and presents challenges including motion blur, appearance change, complex deformation, occlusion, slow motion, and interacting objects. \\[-8pt]

\par{\noindent \textbf{FBMS59~\cite{Ochs14}}}
consists of 59 sequences and 720 annotated frames~(every 20th frame is annotated),
which vary greatly in image resolution.
Sequences involve multiple moving objects, some of which may be static
for periods of time. \\[-8pt]

\par{\noindent \textbf{Moving Camouflaged Animals~(MoCA)~\cite{Lamdouar20}}}
contains 141 HD video sequences, 
depicting 67 kinds of camouflaged animals moving in natural scenes. 
Both temporal and spatial annotations are provided in the form of tight bounding boxes for every 5th frame. 
Using the provided motion labels (locomotion, deformation, static), 
we filter out videos with predominantly no locomotion, 
resulting in 88 video sequences and 4803 frames.

\subsection{Evaluation Metrics}

\par{\noindent \textbf{Segmentation~(Jaccard).}}
\label{sec:metrics}
For DAVIS2016, SegTrackv2 and FBMS59, pixelwise segmentation is provided; 
thus we report the standard metric, region similarity~($\mathcal{J}$), 
computing the mean over the test set. For FBMS59 and SegTrackv2, we follow the common practice \cite{yang_loquercio_2019, Jain17} and combine multiple objects as one single foreground. \\[-8pt]

\par{\noindent \textbf{Localization~(Jaccard \& Success Rate).}}
As the MoCA dataset provides only bounding box annotations,
we evaluate for the detection task 
and report results in the form of detection success rate~\cite{Everingham10,Lin14},
for varying IoU thresholds~($\tau \in \{0.5, 0.6, 0.7, 0.8, 0.9\})$. 


\subsection{Implementation Details}
We evaluate three different approaches for computing optical flow,
namely, PWC-Net~\cite{Sun2018PWC-Net}, RAFT~\cite{Teed20} and ARFlow~\cite{liu2020learning};
the first two are supervised, while the latter is self-supervised. 
We extract the optical flow at the original resolution of the image pairs, 
with the frame gaps $n \in \{-2, -1, 1, 2\}$ for all datasets, 
except for FBMS59, where we use $n \in \{-6, -3, 3, 6\}$ to compensate for small motion. 
To generate inputs to the network for training, the flows are resized to $128 \times 224$~(and scaled accordingly),
converted to 3-channel images with the standard visualization used for optical flow, 
and normalized to $[-1,1]$.

In the iterative binding module~($\Phi_{\text{bind}}(\cdot)$), 
we use two learnable query vectors~(as we consider the case of segmenting a single moving object from the background), and choose $T=5$ iterations~(as explained in Section~\ref{sec:arch}).
We adopt a simple VGG-style network for the CNN encoder and decoder with instance normalization.
We train with a batch size of 64 images and use
the Adam optimizer~\cite{KingmaB14} with an initial learning rate of $5\times 10^{-4}$, 
decreasing every 80k iterations.
The exact architecture description and training schedule can be found in the Supplementary Material.

\section{Results}
\label{sec:results}

\begin{figure*}[!htb]
\begin{floatrow}
\capbtabbox{%
\setlength{\tabcolsep}{5pt}
\begin{tabular}{ccccccc}
\toprule        
 Model   & Flow   & IN   & T & $\mathcal{L}_e$ & $\mathcal{L}_c$ & DAVIS~($\mathcal{J}\uparrow$) \\ \midrule
CIS~\cite{yang_loquercio_2019}   & PWC-Net   &  --   & -- & -- & -- & 59.2 \\ \midrule
Ours-A  & PWC-Net   &  \checkmark    & 5  & \checkmark  & \checkmark  & 63.7 \\ 
\textbf{Ours-B}  & {\bf RAFT}  &  {\bf \checkmark}  & 
{\bf 5} & {\bf \checkmark}  & {\bf \checkmark}  & {\bf 68.3} \\ 
Ours-C  & ARFlow   &  \checkmark    & 5  & \checkmark  & \checkmark  & 53.2 \\  \midrule
Ours-D  & RAFT   &  \checkmark    & 3  & \checkmark  & \checkmark  & 65.8 \\ 
Ours-E  & RAFT   &  \xmark    & 3  & \checkmark  & \checkmark  & 63.3 \\ 
Ours-F  & RAFT   &  \xmark    & 5  & \checkmark  & \checkmark  & 64.5 \\ \midrule
Ours-G  & RAFT   &  \checkmark    & 5  & \xmark  & \xmark  & 48.0 \\ 
Ours-H  & RAFT   &  \checkmark    & 5  & \xmark  & \checkmark  & 60.3 \\ 
Ours-I  & RAFT   &  \checkmark    & 5  & \checkmark  & \xmark  & 51.2 \\ \bottomrule
\end{tabular}
}{%
\caption{\small \textbf{Ablation studies} on flow extraction methods, 
instance normalization~(IN), grouping iterations~($T$), 
entropy regularization~($\mathcal{L}_e$) and set consistency~($\mathcal{L}_c$).}
\label{tab:ablation}}
\hspace{-.46cm}
\ffigbox{%
\includegraphics[width=.49\textwidth]{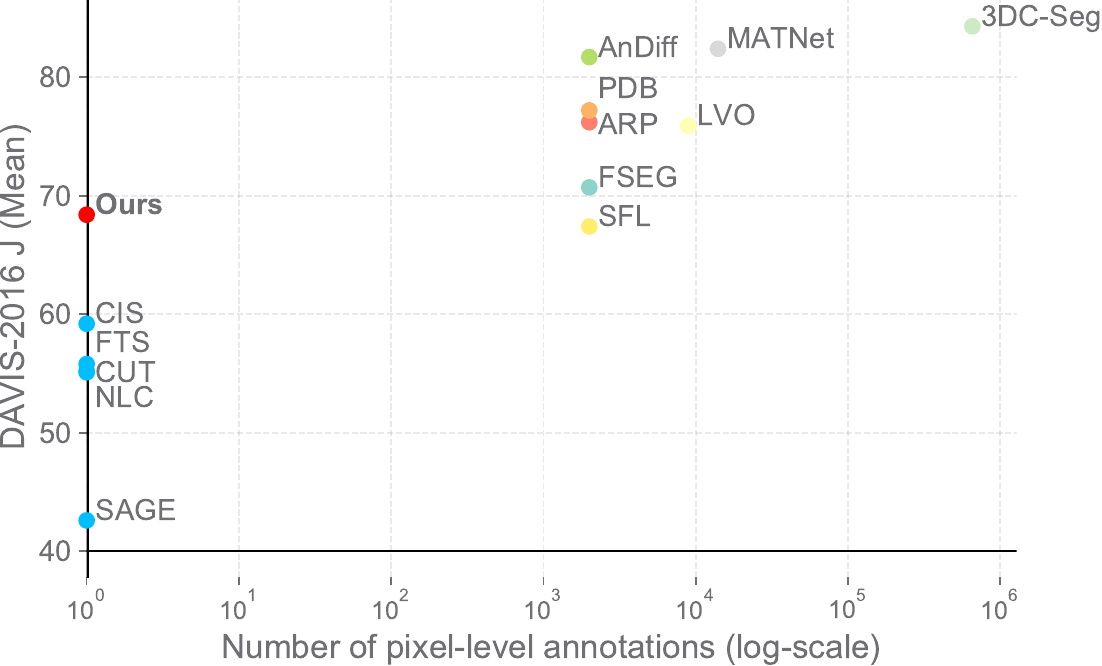}
}{%
\vspace{-12pt}
\caption{ \small
\textbf{Comparison on DAVIS2016.}
Note that, supervised approaches may use ImageNet pretraining~\cite{Deng09},
but here we only count images with pixel-wise annotations.}
\label{fig:comp-graph}
}

\end{floatrow}

\end{figure*}

\begin{table*}[!htb]
\centering
\begin{tabular}{ccccc|cccc}
\toprule        
 Model  & Sup.   & RGB  & Flow   & Res.  & DAVIS16~($\mathcal{J}\uparrow$) 
 & STv2 ~($\mathcal{J}\uparrow$) & FBMS59~($\mathcal{J}\uparrow$)& Runtime~(sec~$\downarrow$) \\ \midrule
SAGE~\cite{sage} & \xmark    & \checkmark  & \checkmark & --  &  42.6  & 57.6 &  {\bf 61.2} & 0.9s \\
NLC~\cite{faktor2014videonlc}  & \xmark    & \checkmark  & \checkmark  & -- &  55.1  & {\bf 67.2} & 51.5   & 11s  \\
CUT~\cite{Keuper15}  & \xmark    & \checkmark  & \checkmark &  -- &  55.2  & 54.3 &  57.2  & 103s \\
FTS~\cite{Papazoglou13} & \xmark    & \checkmark  & \checkmark  & -- &  55.8  & 47.8 & 47.7  & 0.5s  \\
CIS~\cite{yang_loquercio_2019}& \xmark  &\checkmark &\checkmark & $192 \times 384$ &59.2~(71.5) & 45.6 (62.0) & 36.8~(63.5)  & $0.1$s~(11s) \\
{\bf Ours} & {\bf \xmark}  & {\bf \xmark}  & {\bf \checkmark} & {\bf  $128 \times 224$} & {\bf  68.3} & 58.6 & 53.1  & {\bf 0.012s} \\ \midrule
SFL~\cite{Cheng_ICCV_2017}  & \checkmark &  \checkmark & \checkmark  & $854 \times 480$ & 67.4 & --  & -- & 7.9s  \\
FSEG~\cite{Jain17} & \checkmark &  \checkmark & \checkmark  & $854 \times 480$ & 70.7 & 61.4 & 68.4 & --  \\
LVO~\cite{Tokmakov19} & \checkmark &  \checkmark & \checkmark  & -- & 75.9 & 57.3 & 65.1  & --  \\
ARP~\cite{song2018pyramidpdb} & \checkmark &  \checkmark & \checkmark  & -- & 76.2 & 57.2 & 59.8  & 74.5s  \\

COSNet~\cite{Lu_2019_CVPR} & \checkmark &  \checkmark & \xmark  & $473 \times 473$ & 80.5 & -- & 75.6 & --  \\
MATNet~\cite{zhou20} & \checkmark & \checkmark & \checkmark  & $473 \times 473$ & 82.4 &-- & --  & 0.55s  \\  
3DC-Seg~\cite{Mahadevan20BMVC3dc} & \checkmark & \checkmark & \checkmark  & $854 \times 480$ & 84.3 & --& --  & 0.84s  \\  \bottomrule
\end{tabular}
\caption{\small \textbf{Full comparison on moving object segmentation}~(unsupervised video segmentation).
We consider three popular datasets, DAVIS2016, SegTrack-v2 (STv2), and FBMS59.
Models above the horizontal dividing line are trained without using \emph{any} manual annotation,
while models below 
require ground truth annotations at training time.
Numbers in parentheses denote the additional usage of significant post-processing, 
\eg~multi-step flow, multi-crop, temporal smoothing, CRFs.
Runtime excludes optical flow computation.}
\label{tab:main}
\vspace{-5pt}
\end{table*}

\begin{figure*}[tb]
\includegraphics[width=1\textwidth]{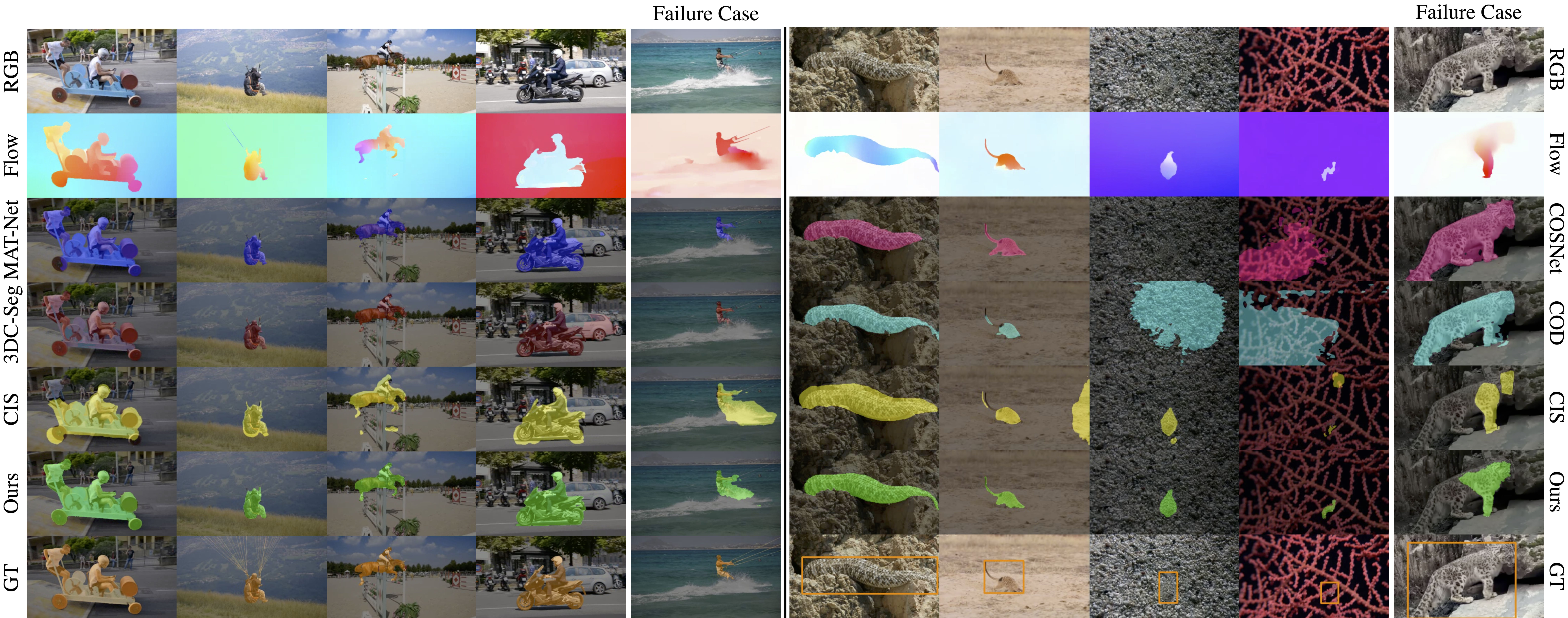}
\centering
\vspace{-5pt}
\caption{\small \textbf{Qualitative results.}
On DAVIS2016 (left), our method is able to segment a variety of challenging objects, often on-par with top supervised approaches. 
On MoCA (right), our model is able to accurately segment well-camouflaged objects even when previous supervised methods fail completely (3rd, 4th columns). We show a failure case (left) where the splash created by the person is incorrectly included in our predicted segment, 
and another failure case (right) where the animal is only partially moving and thus partially segmented.}
\label{fig:results_all}
\end{figure*}

\begin{table*}[tb]
\small
\setlength{\tabcolsep}{6pt}
\centering
\begin{tabular}{l c c c| c | c c c c c c}
\toprule
      & &  &  &  & \multicolumn{6}{c}{Success Rate}                             \\ \toprule
Model & Sup. & RGB & Flow & $\mathcal{J}\uparrow$  & $\tau = 0.5$ & $\tau = 0.6$ & $\tau = 0.7$ & $\tau = 0.8$ & $\tau = 0.9$ & $SR_{mean}$ \\ \midrule
COD~\cite{Lamdouar20}& \checkmark & \xmark & \checkmark  &  44.9  & 0.414 & 0.330 & 0.235 & 0.140 & 0.059 & 0.236 \\ 
COD (two-stream)~\cite{Lamdouar20}& \checkmark & \checkmark & \checkmark & 55.3 & 0.602 & 0.523 & 0.413 & 0.267 & 0.088 & 0.379  \\ 
COSNet~\cite{Lu_2019_CVPR} & \checkmark & \checkmark & \xmark & 50.7  & 0.588 & 0.534 & 0.457 & 0.337 & 0.167 & 0.417 \\
{\bf MATNet}~\cite{zhou20} & {\bf \checkmark} & {\bf \checkmark} &{\bf \checkmark} & {\bf 64.2}  & {\bf 0.712} & {\bf 0.670} & {\bf 0.599} & {\bf 0.492} & {\bf 0.246} & {\bf 0.544} \\ \midrule
CIS  & \xmark & \checkmark& \checkmark&  49.4&  0.556 & 0.463  & 0.329 & 0.176  & 0.030 & 0.311 \\
CIS~(post-processing)  & \xmark & \checkmark & \checkmark&  54.1 &  0.631  & 0.542     & 0.399  & 0.210  & 0.033 & 0.363 \\
{\bf Ours}& {\bf \xmark} & {\bf \xmark} & {\bf \checkmark} &  {\bf 63.4} & {\bf 0.742}  & {\bf 0.654} & {\bf 0.524}  & {\bf 0.351}  & {\bf 0.147} & {\bf 0.484} \\ 
\bottomrule
\end{tabular}
\vspace{-3pt}
\caption{ \small \textbf{Comparison results on MoCA dataset.} 
We report the successful localization rate for various thresholds $\tau$ (see Section~\ref{sec:metrics}).
Both CIS and Ours were pre-trained on DAVIS and finetuned on MoCA in a self-supervised manner.
Our method achieves comparable Jaccard~($\mathcal{J}$) to MATNet~(2nd best model on DAVIS), 
without using RGB inputs and without any manual annotation for training.
}
\label{tab:moca}
\end{table*}

In this section, 
we compare primarily with a top-performing approach trained without manual annotations -- 
Contextual Information Separation~(CIS~\cite{yang_loquercio_2019}).
However,
as the architecture, input resolution, modality and post-processing are all different,
we try our best to conduct the comparison as fairly as possible. 
Note that benchmarks are evaluated at full resolution by simply upsampling the predicted masks.

%

\subsection{Ablation Studies}
\label{sec:ablation}
We conduct all ablation studies on DAVIS2016, and vary one variable each time, as shown in Table \ref{tab:ablation}. \\[-8pt]

\par{\noindent \textbf{Choice of optical flow algorithm.}}
With the same flow extraction method (PWC-Net),
our proposed model~(Ours-A) outperforms CIS by about $4.5$ points on mean Jaccard~($\mathcal{J}$), 
and using improved optical flow~(RAFT) provides further performance gains. 
We therefore use RAFT from hereon. \\[-8pt]

\par{\noindent \textbf{Instance normalization and grouping. }}
We observe two phenomena:
\emph{first}, 
when holding constant on the number of grouping iterations $T$~(3 or 5),
models trained with instance normalization perform consistently better;
\emph{second}, 
iterative grouping with $T=5$ is better than that trained with $T=3$.
However, at $T=8$, the model did not converge in the same number of training steps, 
and thus we do not include it in the table.
For the remainder of the experiments, 
we use instance normalization and $T=5$. \\[-8pt]

\par{\noindent \textbf{Consistency and entropy regularization. }} 
While comparing Ours-B and Ours-I, 
we observe that the performance degrades significantly without the temporal consistency loss,
and that the entropy regularization is also important, as shown by Ours-B and Ours-H.

\subsection{Comparison with State-of-the-art}
We show our results in Table~\ref{tab:main}. 
On DAVIS2016, 
we improve upon the state-of-the-art for unsupervised methods~(CIS) by a large margin ($+9.1\%$).
As shown in Figure~\ref{fig:comp-graph},
despite not using any pixel-level annotations during training,
our method is nearing the performance of supervised models trained on thousands of images.

In addition, 
we argue that, motion segmentation in realistic scenarios, \eg~by predator or prey,
is likely to require fast processing.
Our model operates on small resolution~(potentially sacrificing some accuracy) with over 80fps.
Our method’s efficiency gain mainly comes from two sources: first, our model is a lightweight VGG-style network with only 4.77\textbf{M} parameters; second, we disregard any post-processing used in previous approaches, e.g. averaging the prediction across multiple flow steps, across multiple crops, temporal smoothing, or CRFs, which cost over 10s in total.

For SegTrackv2 and FBMS59, they occasionally include multiple objects in a single video, 
and only a subset of them are moving, 
making it challenging to spot all objects using flow-only input,
but we achieve competitive performance nonetheless. 
We discuss this limitation below. 

\subsection{Camouflage Breaking}
In addition,
we also benchmark the model on camouflaged object detection on MoCA dataset,
where visual cues are often less effective than motion cues.
To compare fairly with CIS~\cite{yang_loquercio_2019},
we use the code and model released by the authors, 
and fine-tune their model on MoCA in a self-supervised manner.
We convert the output segmentation mask into a bounding box by 
drawing a bounding box around the largest connected region in the predicted mask.

We report quantitative results in Table~\ref{tab:moca}
and show qualitative results in Figure~\ref{fig:results_all}.
Our model significantly outperforms CIS~(14\% when allowing no post-processing),
previous supervised approaches \eg~COD~\cite{Lamdouar20}~(18.5\% on Jaccard), 
and even COSNet~\cite{Lu_2019_CVPR}~(among the top supervised approaches on DAVIS).
We conjecture that COSNet's weaker performance is due to its
sole reliance on visual appearance (which is distracting for the MoCA data) rather than using motion inputs.
This is particularly interesting, 
as it clearly indicates that no single information cue is able to do the task perfectly,
echoing the two-stream hypothesis \cite{Goodale92} that both appearance and motion are essential to visual systems.

\subsection{Limitations}
Despite showing remarkable improvements on motion segmentation in accuracy and runtime,
we note the following limitations of the proposed approach~(shown in Figure~\ref{fig:results_all}) and treat them as future work:
\emph{first},
the existing benchmarks are mostly limited to motion segmentation into foreground and background, 
thus, we choose to use two slots in this paper;
however, in real scenarios, videos may contain multiple independently moving objects, 
which the current model will assign to a single layer.
It may be desirable to further separate these objects into different layers. 
\emph{Second},
we have only explored motion~(optical flow) as input, 
which significantly limits the model in segmenting objects when flow is uninformative or incomplete~(as in Figure~\ref{fig:results_all}, right); 
however, the self-supervised video object segmentation objective is applicable also to a two-stream approach, and so RGB could be incorporated. 
\emph{Third}, the current method may fail when optical flow is noisy or low-quality~(Figure~\ref{fig:results_all}, left); 
jointly optimizing flow and segmentation is a possible way forward in this case.

\section{Conclusion}
In this paper, 
we present a self-supervised model for motion segmentation.
The algorithm takes only flow as input, and is trained without any manual annotation, 
surpassing previous self-supervised methods on public benchmarks such as DAVIS2016, 
narrowing the gap with supervised methods.
On the more challenging camouflage dataset~(MoCA), 
our model actually compares favourably to the top approaches in video object segmentation that are trained with heavy supervision.
As computation power grows and more high quality videos become available, 
we believe that self-supervised learning algorithms can serve as a strong competitor to the supervised counterparts for their scalability and generalizability.

\vspace{-5pt}
\section{Acknowledgements}
\vspace{-5pt}
This research is supported by Google-DeepMind Studentship, 
UK EPSRC CDT in AIMS, 
Schlumberger Studentship, and the UK EPSRC Programme Grant Visual AI (EP/T028572/1).


{\small
\bibliographystyle{ieee_fullname}
\bibliography{longstrings,egbib,vgg_local,vgg_other}
}
\newpage
\appendix
\section{Training Details}

In this section, we include the details for reproducing our results, 
for example, architectures, pseudo-codes and hyper-parameters.

\subsection{Encoder \& Decoder}
The backbone of network architecture for the model are shown in Table~\ref{tab:cnn},
we refer the readers to the pseudocode for iterative binding module.

\begin{table}[!htb]
\small
\begin{tabular}{c|l|l|l}
\multicolumn{1}{l|}{} & stage & operation & output sizes \\ \cline{2-4} 
\multicolumn{1}{l|}{} & input & -- & $ 3 \times 128 \times 224$ \\ \hline
\multirow{5}{*}{\rotatebox[origin=c]{90}{Encoder}} & conv1 & {[}$5 \times 5, 64${]} $\times$ 2 & $ 64 \times 128 \times 224$ \\
 & mp1 & maxpool, stride = $2$ & $ 64 \times 64 \times 112$ \\
 & conv2 & {[}$5 \times 5, 128${]} $\times$ 2 & $ 128 \times 64 \times 112$ \\
 & mp2 & maxpool, stride = $2$  & $ 128 \times 32 \times 56$ \\
 & conv3 & {[}$5 \times 5, 256${]} $\times$ 2 & $ 256 \times 32 \times 56$ \\ \midrule


\multirow{5}{*}{\rotatebox[origin=c]{90}{Decoder}} & conv$^T$1 & $5 \times 5, 64, \text{stride} = 2$ & $ 64 \times 2 \times 32 \times 56$ \\
 & conv$^T$2 & $5 \times 5, 64, \text{stride} = 2$ & $ 64 \times 2 \times 64 \times 112$ \\
 & conv$^T$3 & $5 \times 5, 64, \text{stride} = 2$ & $ 64 \times 2 \times 128 \times 224$ \\
 & outconv & \begin{tabular}[c]{@{}l@{}}$5 \times 5, 64$,\\ $5 \times 5, 4$\end{tabular} & $ 4 \times 2 \times 128 \times 224$
\end{tabular}
\caption{Network architecture. All convolutions have padding 2 to preserve spatial resolution, and are followed by instance normalization and ReLU activation, except the final layer. The details of the iterative binding module is in Figure \ref{fig:code}.}
\label{tab:cnn}
\end{table}

\subsection{Iterative Binding Module}
\begin{figure}[!htb]
\includegraphics[width=\textwidth]{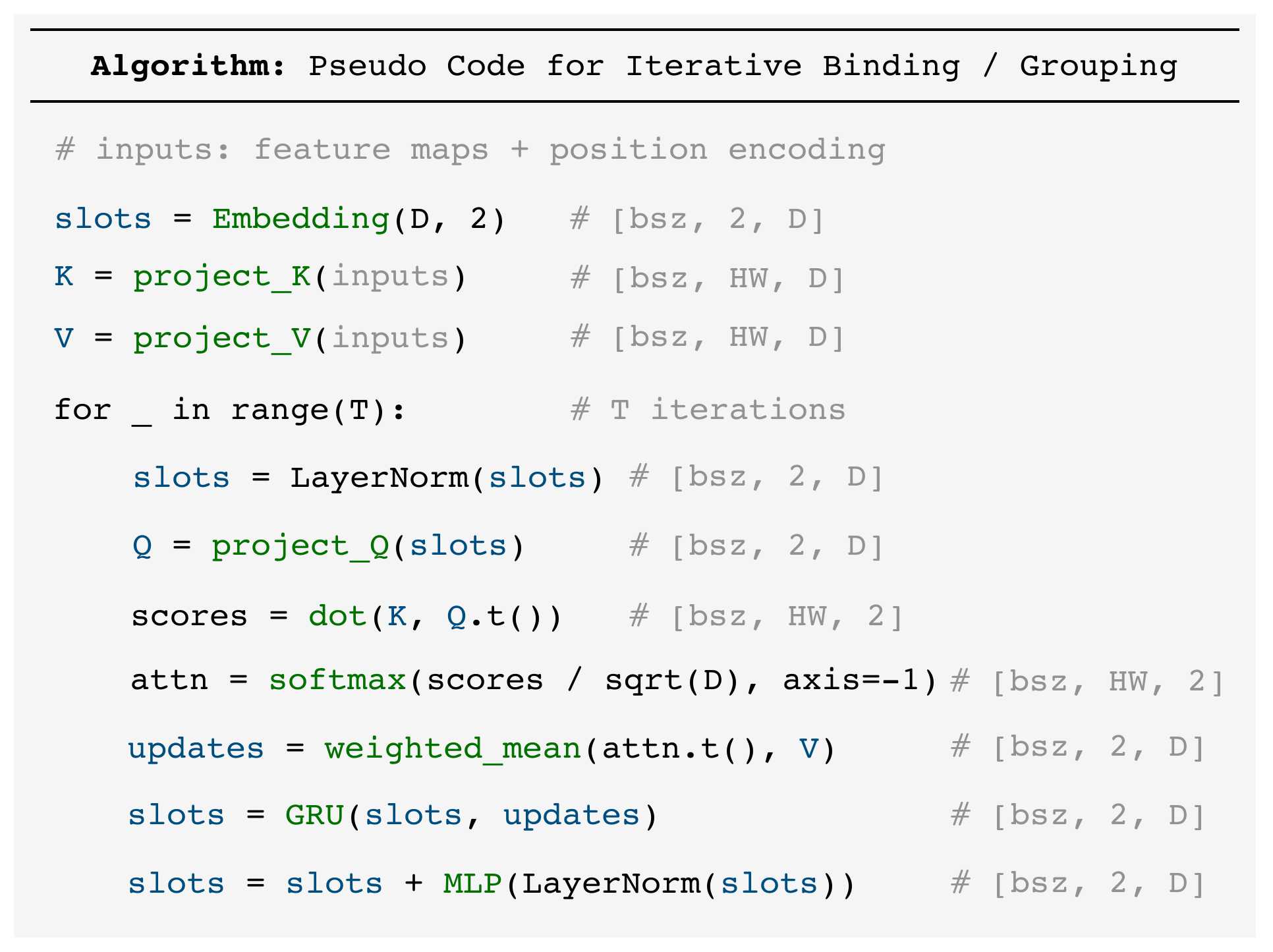}
\centering
\caption{Pseudocode for iterative binding. The linear projections for key (K) and value (V) have 256 dimensions. The MLP has two layers, both with 256 dimensions, with ReLU in between. }
\label{fig:code}
\end{figure}

The pseudocode for the iterative binding module is shown in Figure \ref{fig:code}. 
The full code is available in the submitted source code, and will be made publicly available.

\subsection{Hyperparameters}
For all datasets, we train with batch size of 64, 
although note that this corresponds to 32 \textit{pairs} of optical flow in order to train with consistency loss. 
Our initial learning rate is 5e-4, with first 200 steps being warmup,
and decays by half every 8e4 iterations. 
During this decay, the scale for entropy and consistency loss is also increased by a factor of 5,
gradually encouraging the predicted alpha channel to be binary.
We train the algorithm for about 300k iterations.

\section{MoCA dataset curation}
The MoCA dataset
contains 141 high-definition video sequences, 
with an average duration of 11 seconds. These sequences were collected from YouTube with resolution $720 \times 1280$, and sampled at 24 fps, resulting in $37$K frames
depicting 67 kinds of camouflaged animals moving in natural scenes. 
Both temporal and spatial annotations are provided in the form of tight bounding boxes on every 5th frame. 
We use a modified version of this dataset in order to make it more suitable for segmentation tasks. 
We outline the process and rationale below.
\begin{itemize}
    \vspace{-0.2cm}\item We crop away the channel logos and empty border spaces, and resize the low-resolution images to the same resolution as the other images in the dataset (at $720 \times 1280$).
    We then adjust the ground-truth annotations accordingly.
    \vspace{-0.2cm}\item The original authors resampled all videos to 24 fps even when some original videos have less, causing some consecutive frames to be identical. To alleviate this, we sample every 3 frames from the original dataset, up to 100 frames per video. 
    \vspace{-0.2cm}\item For annotations, we use linear interpolation to generate the missing frames' bounding boxes, 
    resulting in a dense frame-wise annotation. 
    \vspace{-0.2cm}\item The authors provided the motion labels for each annotated frame (locomotion, deformation, static), so we filter away videos with predominantly no locomotion. 
    \vspace{-0.2cm}\item We also further discard videos that contain large amount of frames where the motion does not belong to the primary object. 
\end{itemize}
This eventually results in 88 video sequences and 4803 frames, 
which we will release for fair comparison. \\[-10pt]

\section{Results breakdown}
The main evaluation metric used in this paper is the Jaccard score ($\mathcal{J}$), which is the intersection-over-union between the predicted and ground-truth masks.
In line with previous works, 
we show the per-category results breakdown for our model on DAVIS2016, SegTrackv2, FBMS59, and MoCA in Tables \ref{tab:stv2}-\ref{tab:moca2}.
Note that, since we focus on both speed and accuracy,
the predictions are only of $128 \times 224$ pixels, 
and we directly upsample this prediction to the original resolution and compare with the groundtruth. 
\section{Qualitative results}
\begin{figure*}[!htb]
\includegraphics[width=\textwidth]{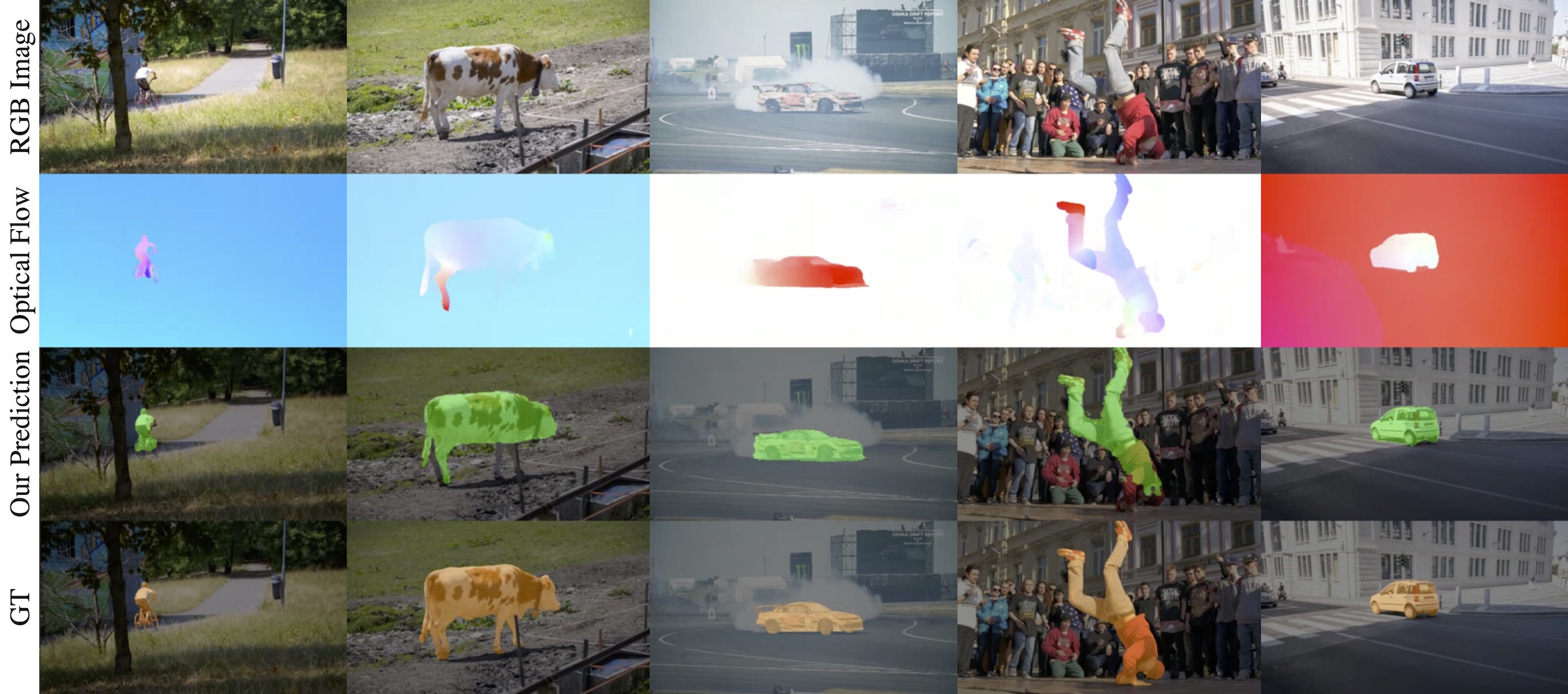}
\includegraphics[width=\textwidth]{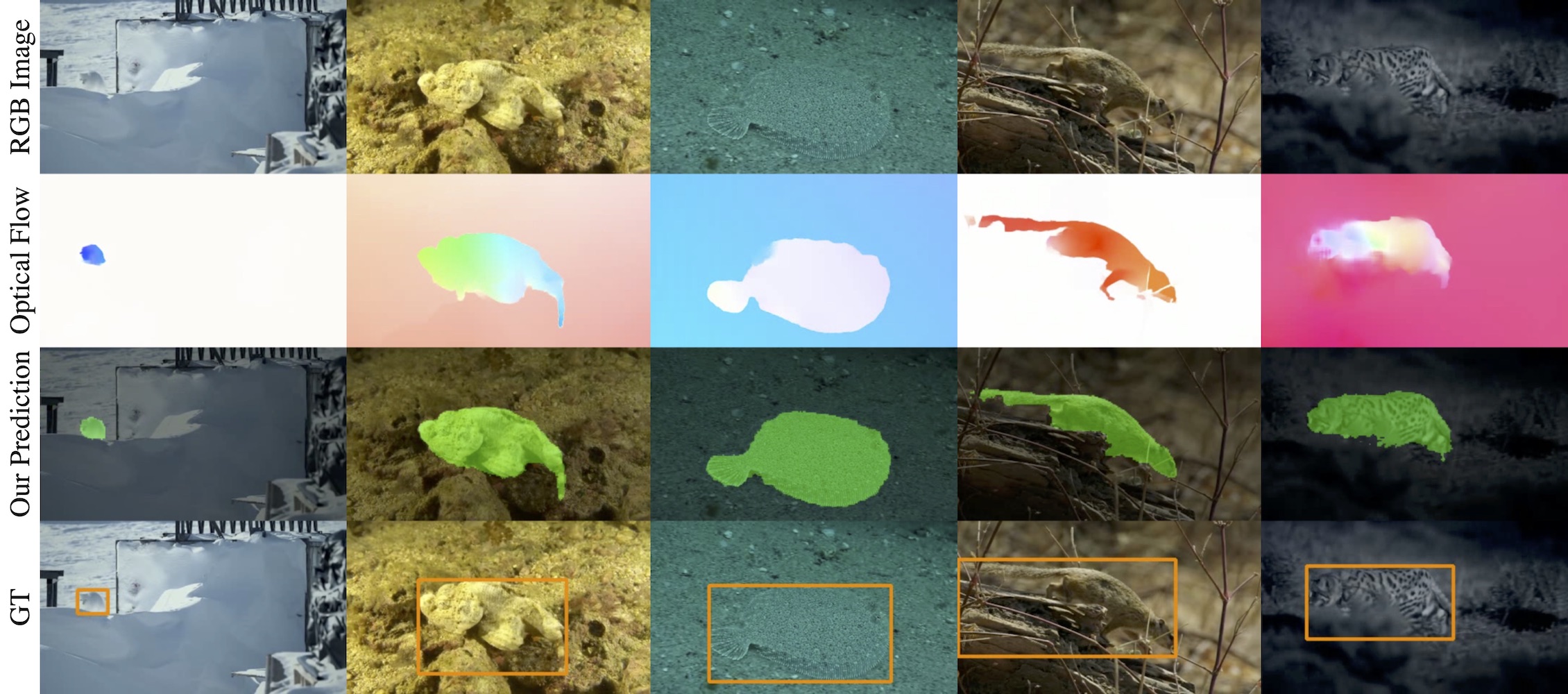}
\centering
\caption{Qualitative results on DAVIS2016 and MoCA, respectively.}
\label{fig:supp}
\end{figure*}

\begin{figure*}[!htb]
\includegraphics[width=\textwidth]{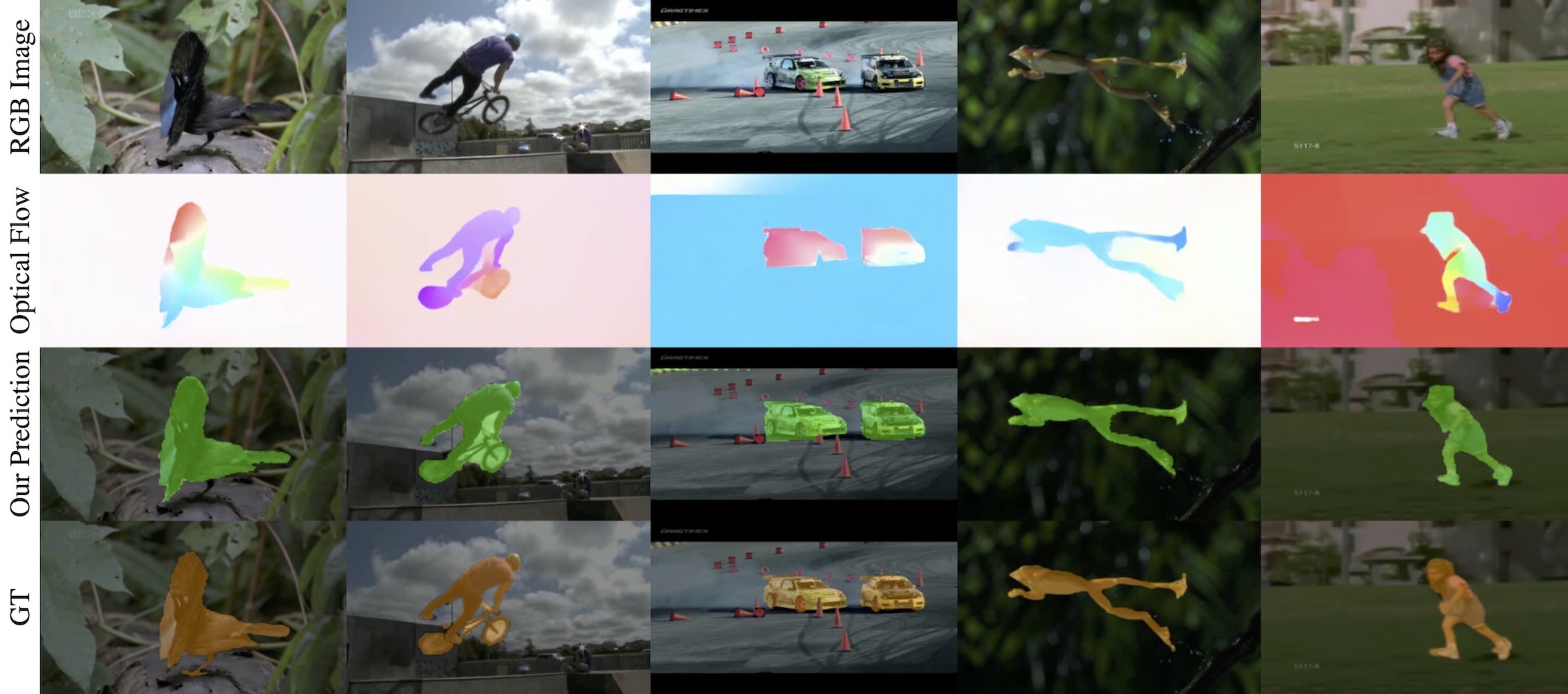}
\includegraphics[width=\textwidth]{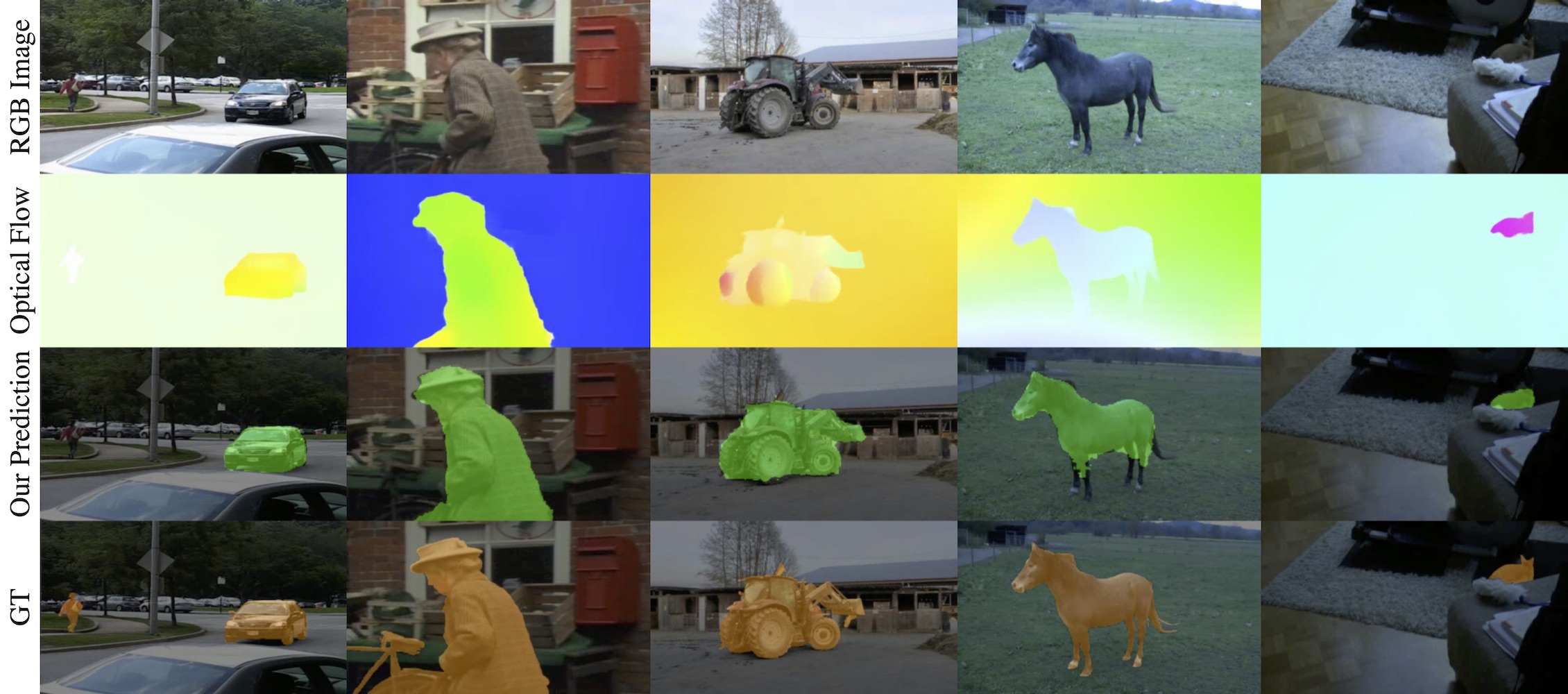}
\centering
\caption{Qualitative results on SegTrackv2 and FBMS59, respectively.}
\label{fig:supp2}
\end{figure*}

We show more qualitative results in Figures \ref{fig:supp} and \ref{fig:supp2}.

\begin{table}[!htb]
\centering
\footnotesize
\begin{tabular}{c|c}

Sequence&$\mathcal{J}$(M)\\ \hline
bird of paradise & 0.791 \\ 
birdfall & 0.300 \\ 
bmx & 0.609 \\ 
cheetah & 0.370 \\ 
drift & 0.797 \\ 
frog & 0.733 \\ 
girl & 0.746 \\ 
hummingbird & 0.506 \\ 
monkey & 0.751 \\ 
monkeydog & 0.133 \\ 
parachute & 0.914 \\ 
penguin & 0.697 \\ 
soldier & 0.741 \\ 
worm & 0.326 \\ \hline
seq avg & 0.601 \\ 
frames avg & 0.586 \\ 
\end{tabular}
\caption{Sequence-wise results on SegTrackv2.}
\label{tab:stv2}
\end{table}

\begin{table}[!htb]
\footnotesize
\centering
\begin{tabular}{c|ccc|ccc}

Sequence&$\mathcal{J}$(M)&$\mathcal{J}$(R)&$\mathcal{J}$(D)&$\mathcal{F}$(M)&$\mathcal{F}$(R)&$\mathcal{F}$(D)\\ \hline
bear& 0.766& 1.000&-4.5& 0.640& 0.914&-1.9\\
blackswan& 0.795& 1.000& 5.9& 0.658& 0.980& 7.3\\
bmx-bumps& 0.373& 0.114& 8.3& 0.260& 0.011& 10.8\\
bmx-trees& 0.744& 1.000& 14.5& 0.666& 0.909& 13.9\\
boat& 0.602& 0.787& 44.5& 0.724& 0.936& 22.4\\
breakdance& 0.792& 0.990&-2.4& 0.814& 1.000& 10.8\\
breakdance-flare& 0.826& 1.000& 10.2& 0.771& 1.000& 16.7\\
bus& 0.391& 0.000&-4.4& 0.248& 0.000&-5.0\\
camel& 0.540& 0.679& 9.3& 0.667& 0.923&-9.8\\
car-roundabout& 0.479& 0.500& 1.2& 0.404& 0.146&-10.6\\
car-shadow& 0.879& 1.000&-0.2& 0.818& 1.000&-3.3\\
car-turn& 0.492& 0.537& 3.4& 0.560& 0.683& 6.8\\
cows& 0.690& 0.898&-5.0& 0.653& 0.852& 4.4\\
dance-jump& 0.815& 1.000&-4.8& 0.683& 1.000& 5.9\\
dance-twirl& 0.739& 0.960& 7.4& 0.803& 0.900&28.5\\
dog& 0.701& 0.789& 2.9& 0.477& 0.526& 9.3\\
dog-agility& 0.810& 1.000& 7.7& 0.858& 1.000& 4.0\\
drift-chicane& 0.777& 1.000& 7.3& 0.611& 0.646& 26.3\\
drift-straight& 0.864& 1.000&-5.2& 0.685& 1.000&-6.8\\
drift-turn& 0.593& 0.654& 27.7& 0.220& 0.000& 20.4\\ \hline
Average& 0.683& 0.795& 6.2 & 0.611& 0.721& 7.5  \\ 
\end{tabular}
\caption{Full results on DAVIS2016. J refers to the Jaccard score, while the F-measure refers to the contour accuracy. M, R, and D refers to mean, recall and decay respectively.}
\label{tab:davis}
\end{table}

\begin{table}[!htb]
\centering
\footnotesize
\begin{tabular}{c|c}

Sequence&$\mathcal{J}$(M)\\ \hline
camel01 & 0.281 \\ 
cars1 & 0.846 \\ 
cars10 & 0.322 \\ 
cars4 & 0.826 \\ 
cars5 & 0.842 \\ 
cats01 & 0.672 \\ 
cats03 & 0.640 \\ 
cats06 & 0.362 \\ 
dogs01 & 0.629 \\ 
dogs02 & 0.636 \\ 
farm01 & 0.816 \\ 
giraffes01 & 0.322 \\ 
goats01 & 0.375 \\ 
horses02 & 0.628 \\ 
horses04 & 0.566 \\ 
horses05 & 0.334 \\ 
lion01 & 0.399 \\ 
marple12 & 0.680 \\ 
marple2 & 0.750 \\ 
marple4 & 0.799 \\ 
marple6 & 0.450 \\ 
marple7 & 0.567 \\ 
marple9 & 0.537 \\ 
people03 & 0.598 \\ 
people1 & 0.761 \\ 
people2 & 0.842 \\ 
rabbits02 & 0.415 \\  
rabbits03 & 0.319 \\ 
rabbits04 & 0.400 \\ 
tennis & 0.561 \\ \hline
seq avg & 0.573 \\ 
frames avg & 0.531 \\ 
\end{tabular}
\caption{Sequence-wise results on FBMS59.}
\label{tab:fbms}
\end{table}

\newpage

\begin{table*}[]
\begin{tabular}{l|c|cccccc}
sequence & $\mathcal{J}$ & $\tau_{0.5}$ & $\tau_{0.6}$ & $\tau_{0.7}$ & $\tau_{0.8}$ & $\tau_{0.9}$ & $avg$ \\ \hline
arabian\_horn\_viper & 0.709 & 0.990 & 0.909 & 0.586 & 0.091 & 0.000 & 0.515 \\
arctic\_fox & 0.381 & 0.404 & 0.383 & 0.298 & 0.191 & 0.000 & 0.255 \\
arctic\_fox\_1 & 0.879 & 0.913 & 0.913 & 0.913 & 0.913 & 0.652 & 0.861 \\
arctic\_wolf\_0 & 0.705 & 0.929 & 0.859 & 0.596 & 0.202 & 0.051 & 0.527 \\ 
arctic\_wolf\_1 & 0.712 & 0.795 & 0.795 & 0.795 & 0.744 & 0.256 & 0.677 \\
bear & 0.508 & 0.611 & 0.453 & 0.221 & 0.074 & 0.000 & 0.272 \\
black\_cat\_0 & 0.499 & 0.556 & 0.476 & 0.302 & 0.048 & 0.000 & 0.276 \\
black\_cat\_1 & 0.086 & 0.030 & 0.020 & 0.010 & 0.000 & 0.000 & 0.012 \\
crab & 0.594 & 0.800 & 0.400 & 0.200 & 0.000 & 0.000 & 0.280 \\
crab\_1 & 0.288 & 0.309 & 0.200 & 0.073 & 0.000 & 0.000 & 0.116 \\
cuttlefish\_0 & 0.222 & 0.194 & 0.032 & 0.000 & 0.000 & 0.000 & 0.045 \\
cuttlefish\_1 & 0.034 & 0.043 & 0.000 & 0.000 & 0.000 & 0.000 & 0.009 \\
cuttlefish\_4 & 0.655 & 1.000 & 0.846 & 0.231 & 0.000 & 0.000 & 0.415 \\
cuttlefish\_5 & 0.724 & 0.870 & 0.870 & 0.783 & 0.304 & 0.043 & 0.574 \\
dead\_leaf\_butterfly\_1 & 0.784 & 0.913 & 0.783 & 0.739 & 0.696 & 0.304 & 0.687 \\
desert\_fox & 0.470 & 0.362 & 0.234 & 0.191 & 0.149 & 0.106 & 0.209 \\
devil\_scorpionfish & 0.938 & 1.000 & 1.000 & 1.000 & 0.913 & 0.826 & 0.948 \\
devil\_scorpionfish\_1 & 0.913 & 1.000 & 1.000 & 1.000 & 1.000 & 0.565 & 0.913 \\
devil\_scorpionfish\_2 & 0.857 & 0.968 & 0.903 & 0.871 & 0.806 & 0.548 & 0.819 \\
egyptian\_nightjar & 0.765 & 0.905 & 0.842 & 0.789 & 0.611 & 0.158 & 0.661 \\
elephant & 0.728 & 0.783 & 0.652 & 0.609 & 0.565 & 0.348 & 0.591 \\
flatfish\_0 & 0.575 & 0.677 & 0.657 & 0.636 & 0.485 & 0.141 & 0.519 \\
flatfish\_1 & 0.682 & 0.848 & 0.835 & 0.722 & 0.354 & 0.051 & 0.562 \\
flatfish\_2 & 0.765 & 0.839 & 0.774 & 0.774 & 0.742 & 0.645 & 0.755 \\
flatfish\_4 & 0.697 & 0.958 & 0.895 & 0.568 & 0.126 & 0.000 & 0.509 \\
flounder & 0.896 & 1.000 & 1.000 & 0.986 & 0.986 & 0.437 & 0.882 \\
flounder\_3 & 0.505 & 0.429 & 0.286 & 0.143 & 0.143 & 0.000 & 0.200 \\
flounder\_4 & 0.767 & 0.949 & 0.897 & 0.769 & 0.487 & 0.128 & 0.646 \\
flounder\_5 & 0.681 & 0.797 & 0.747 & 0.696 & 0.468 & 0.165 & 0.575 \\
flounder\_6 & 0.683 & 0.768 & 0.677 & 0.616 & 0.465 & 0.263 & 0.558 \\
flounder\_7 & 0.719 & 0.930 & 0.845 & 0.662 & 0.254 & 0.028 & 0.544 \\
flounder\_8 & 0.707 & 0.925 & 0.774 & 0.645 & 0.409 & 0.000 & 0.551 \\
flounder\_9 & 0.636 & 0.821 & 0.718 & 0.436 & 0.231 & 0.026 & 0.446 \\
fossa & 0.280 & 0.143 & 0.000 & 0.000 & 0.000 & 0.000 & 0.029 \\
goat\_0 & 0.589 & 0.707 & 0.636 & 0.414 & 0.212 & 0.020 & 0.398 \\
goat\_1 & 0.744 & 0.930 & 0.930 & 0.704 & 0.380 & 0.085 & 0.606 \\
groundhog & 0.525 & 0.646 & 0.525 & 0.374 & 0.162 & 0.010 & 0.343 \\
hedgehog\_0 & 0.329 & 0.298 & 0.170 & 0.085 & 0.021 & 0.021 & 0.119 \\
hedgehog\_1 & 0.471 & 0.533 & 0.400 & 0.333 & 0.067 & 0.067 & 0.280 \\
hedgehog\_2 & 0.771 & 0.800 & 0.733 & 0.733 & 0.600 & 0.267 & 0.627 \\
hedgehog\_3 & 0.486 & 0.564 & 0.359 & 0.282 & 0.128 & 0.026 & 0.272 \\
hermit\_crab & 0.674 & 0.806 & 0.806 & 0.677 & 0.419 & 0.065 & 0.555 \\
ibex & 0.390 & 0.513 & 0.359 & 0.128 & 0.000 & 0.000 & 0.200 \\
jerboa & 0.555 & 0.739 & 0.435 & 0.348 & 0.130 & 0.000 & 0.330
\end{tabular}
\label{tab:moca2}
\end{table*}

\begin{table*}[]
\begin{tabular}{l|c|cccccc}
sequence & $\mathcal{J}$ & $\tau_{0.5}$ & $\tau_{0.6}$ & $\tau_{0.7}$ & $\tau_{0.8}$ & $\tau_{0.9}$ & $avg$ \\ \hline
jerboa\_1 & 0.450 & 0.452 & 0.323 & 0.226 & 0.097 & 0.000 & 0.219 \\
lichen\_katydid & 0.502 & 0.455 & 0.323 & 0.162 & 0.020 & 0.010 & 0.194 \\
lion\_cub\_0 & 0.728 & 0.972 & 0.873 & 0.549 & 0.282 & 0.127 & 0.561 \\
lion\_cub\_1 & 0.571 & 0.687 & 0.556 & 0.354 & 0.141 & 0.010 & 0.349 \\ 
lion\_cub\_3 & 0.179 & 0.182 & 0.141 & 0.081 & 0.051 & 0.000 & 0.091 \\ 
lioness & 0.423 & 0.419 & 0.129 & 0.000 & 0.000 & 0.000 & 0.110 \\
marine\_iguana & 0.376 & 0.217 & 0.130 & 0.000 & 0.000 & 0.000 & 0.070 \\
markhor & 0.808 & 0.909 & 0.855 & 0.800 & 0.709 & 0.364 & 0.727 \\
meerkat & 0.778 & 0.871 & 0.774 & 0.742 & 0.710 & 0.323 & 0.684 \\
mountain\_goat & 0.742 & 1.000 & 0.968 & 0.710 & 0.226 & 0.065 & 0.594 \\
nile\_monitor\_1 & 0.524 & 0.616 & 0.434 & 0.283 & 0.121 & 0.000 & 0.291 \\
octopus & 0.637 & 0.838 & 0.687 & 0.273 & 0.131 & 0.020 & 0.390 \\
octopus\_1 & 0.411 & 0.242 & 0.091 & 0.061 & 0.061 & 0.010 & 0.093 \\
peacock\_flounder\_0 & 0.884 & 0.931 & 0.931 & 0.931 & 0.931 & 0.701 & 0.885 \\
peacock\_flounder\_1 & 0.812 & 0.970 & 0.929 & 0.859 & 0.667 & 0.222 & 0.729 \\
peacock\_flounder\_2 & 0.873 & 1.000 & 1.000 & 0.989 & 0.926 & 0.368 & 0.857 \\
polar\_bear\_0 & 0.622 & 0.845 & 0.676 & 0.352 & 0.141 & 0.000 & 0.403 \\
polar\_bear\_1 & 0.487 & 0.603 & 0.556 & 0.476 & 0.175 & 0.016 & 0.365 \\
polar\_bear\_2 & 0.792 & 0.949 & 0.846 & 0.744 & 0.641 & 0.308 & 0.697 \\
pygmy\_seahorse\_2 & 0.478 & 0.582 & 0.364 & 0.255 & 0.036 & 0.000 & 0.247 \\
pygmy\_seahorse\_4 & 0.654 & 0.935 & 0.839 & 0.516 & 0.000 & 0.000 & 0.458 \\
rodent\_x & 0.738 & 0.870 & 0.826 & 0.696 & 0.435 & 0.174 & 0.600 \\
scorpionfish\_0 & 0.622 & 0.761 & 0.648 & 0.521 & 0.408 & 0.127 & 0.493 \\
scorpionfish\_1 & 0.616 & 0.681 & 0.574 & 0.426 & 0.319 & 0.128 & 0.426 \\
scorpionfish\_2 & 0.842 & 0.962 & 0.949 & 0.924 & 0.823 & 0.380 & 0.808 \\
scorpionfish\_3 & 0.804 & 0.975 & 0.861 & 0.785 & 0.595 & 0.354 & 0.714 \\
scorpionfish\_4 & 0.800 & 1.000 & 0.949 & 0.821 & 0.513 & 0.231 & 0.703 \\
scorpionfish\_5 & 0.899 & 1.000 & 1.000 & 1.000 & 0.957 & 0.478 & 0.887 \\
seal\_1 & 0.865 & 1.000 & 0.913 & 0.870 & 0.870 & 0.652 & 0.861 \\
seal\_2 & 0.676 & 0.800 & 0.655 & 0.455 & 0.291 & 0.145 & 0.469 \\
seal\_3 & 0.445 & 0.400 & 0.200 & 0.067 & 0.000 & 0.000 & 0.133 \\
shrimp & 0.772 & 0.933 & 0.867 & 0.733 & 0.667 & 0.133 & 0.667 \\
snow\_leopard\_0 & 0.760 & 1.000 & 0.949 & 0.692 & 0.359 & 0.077 & 0.615 \\
snow\_leopard\_1 & 0.772 & 0.826 & 0.696 & 0.652 & 0.652 & 0.522 & 0.670 \\
snow\_leopard\_2 & 0.883 & 1.000 & 0.989 & 0.968 & 0.863 & 0.526 & 0.869 \\
snow\_leopard\_3 & 0.608 & 0.778 & 0.556 & 0.417 & 0.333 & 0.083 & 0.433 \\
snow\_leopard\_6 & 0.816 & 0.830 & 0.787 & 0.787 & 0.723 & 0.660 & 0.757 \\
snow\_leopard\_7 & 0.679 & 0.871 & 0.806 & 0.581 & 0.258 & 0.000 & 0.503 \\
snow\_leopard\_8 & 0.556 & 0.702 & 0.596 & 0.447 & 0.191 & 0.000 & 0.387 \\
snowy\_owl\_0 & 0.564 & 0.532 & 0.340 & 0.298 & 0.128 & 0.000 & 0.260 \\
spider\_tailed\_horned\_viper\_0 & 0.473 & 0.400 & 0.333 & 0.267 & 0.067 & 0.067 & 0.227 \\
spider\_tailed\_horned\_viper\_1 & 0.558 & 0.620 & 0.423 & 0.310 & 0.254 & 0.056 & 0.332 \\
spider\_tailed\_horned\_viper\_2 & 0.848 & 0.964 & 0.945 & 0.909 & 0.782 & 0.564 & 0.833 \\
spider\_tailed\_horned\_viper\_3 & 0.860 & 1.000 & 1.000 & 1.000 & 0.677 & 0.387 & 0.813 \\ \hline
seq avg & 0.634 &	0.734 &	0.640 &	0.522 &	0.361 &	0.166 &	0.485 \\
frames avg & 0.634 & 0.742 & 0.654 & 0.524 & 0.351 & 0.147 & 0.484
\end{tabular}
\caption{Results breakdown for MoCA.}
\label{tab:moca2}
\end{table*}

\newpage

\newpage

\end{document}